\documentclass[journal]{IEEEtran}

\usepackage{xcolor,soul,framed} 

\colorlet{shadecolor}{yellow}
\usepackage[pdftex]{graphicx}
\graphicspath{{../pdf/}{../jpeg/}}
\DeclareGraphicsExtensions{.pdf,.jpeg,.png}

\usepackage[cmex10]{amsmath}
\usepackage{array}
\usepackage{mdwmath}
\usepackage{mdwtab}
\usepackage{eqparbox}
\usepackage{url}
\usepackage{hyperref}
\hypersetup{
  colorlinks=false,
  pdfborder={0 0 0},  
}

\usepackage{float}
\begin{document}
\bstctlcite{IEEEexample:BSTcontrol}
    \title{Securing Social Media Against Deepfakes using Identity, Behavioral, and Geometric Signatures
}


  \author{Muhammad~Umar~Farooq,
      Awais~Khan,
      Ijaz~Ul~Haq,
      and Khalid~Mahmood~Malik,~\IEEEmembership{Senior~Member,~IEEE}%

  \thanks{Manuscript received XXX. This work is supported by the National Science Foundation (NSF) under award numbers 2329858 and 2231619 and Michigan Translational Research and Commercialization (MTRAC), Advanced Computing Technologies (ACT) award number 292883. This material is based upon work supported by the National Science Foundation under Grant No. 2409577}
  \thanks{Muhammad Umar Farooq, Awais Khan, Ijaz Ul Haq and Khalid Mahmood Malik are with  College of Innovation and Technology, University of Michigan-Flint, Flint, MI 48502, USA (e-mail: mufarooq@umich.edu, mawais@umich.edu, Ijazhaq@umich.edu and drmalik@umich.edu).}}

\markboth{Journal of \LaTeX\ Class Files,~Vol.~14, No.~8, December~2024 
}{}

\maketitle

\begin{abstract}

Trust in social media is a growing concern due to its ability to influence significant societal changes. However, this space is increasingly compromised by various types of deepfake multimedia, which undermine the authenticity of shared content. Although substantial efforts have been made to address the challenge of deepfake content, existing detection techniques face a major limitation in generalization: they tend to perform well only on specific types of deepfakes they were trained on.This dependency on recognizing specific deepfake artifacts makes current methods vulnerable when applied to unseen or varied deepfakes, thereby compromising their performance in real-world applications such as social media platforms. To address the generalizability of deepfake detection, there is a need for a holistic approach that can capture a broader range of facial attributes and manipulations beyond isolated artifacts. To address this, we propose a novel deepfake detection framework featuring an effective feature descriptor that integrates Deep identity, Behavioral, and Geometric (DBaG) signatures, along with a classifier named DBaGNet. Specifically, the DBaGNet classifier utilizes the extracted DBaG signatures, leveraging a triplet loss objective to enhance generalized representation learning for improved classification. Specifically, the DBaGNet classifier utilizes the extracted DBaG signatures and applies a triplet loss objective to enhance generalized representation learning for improved classification. The comprehensive DBaG signatures captures both facial geometry inconsistencies and behavioral cues, enhancing the detection of diverse deepfake types and improving generalization. To test the effectiveness and generalizability of our proposed approach, we conduct extensive experiments using six benchmark deepfake datasets: WLDR, CelebDF, DFDC, FaceForensics++, DFD, and NVFAIR. Specifically, to ensure the effectiveness of our approach, we perform cross-dataset evaluations, and the results demonstrate significant performance gains over several state-of-the-art methods.

\end{abstract}

\begin{IEEEkeywords}
Deepfake Detection, Multimedia Forensics, Behavioral Biometrics, face forgery detection, DBaG
\end{IEEEkeywords}

%
\IEEEpeerreviewmaketitle


\section{Introduction}
Over the past decade, internet traffic has seen a significant shift from text-based information to multimedia files, driven by the rise of large-scale social multimedia platforms \cite{zhao2009behavior}. While these platforms enrich and facilitate the sharing of everyday experiences, they also pose a growing threat: the spread of fake multimedia (e.g., images, videos, and audio) that contains misleading or fabricated content. Advances in video generation technologies have played a crucial role in the rise of such forged media through various forms of manipulation.
\begin{figure}[t]
  \begin{center}
	\includegraphics[width=3.5in]{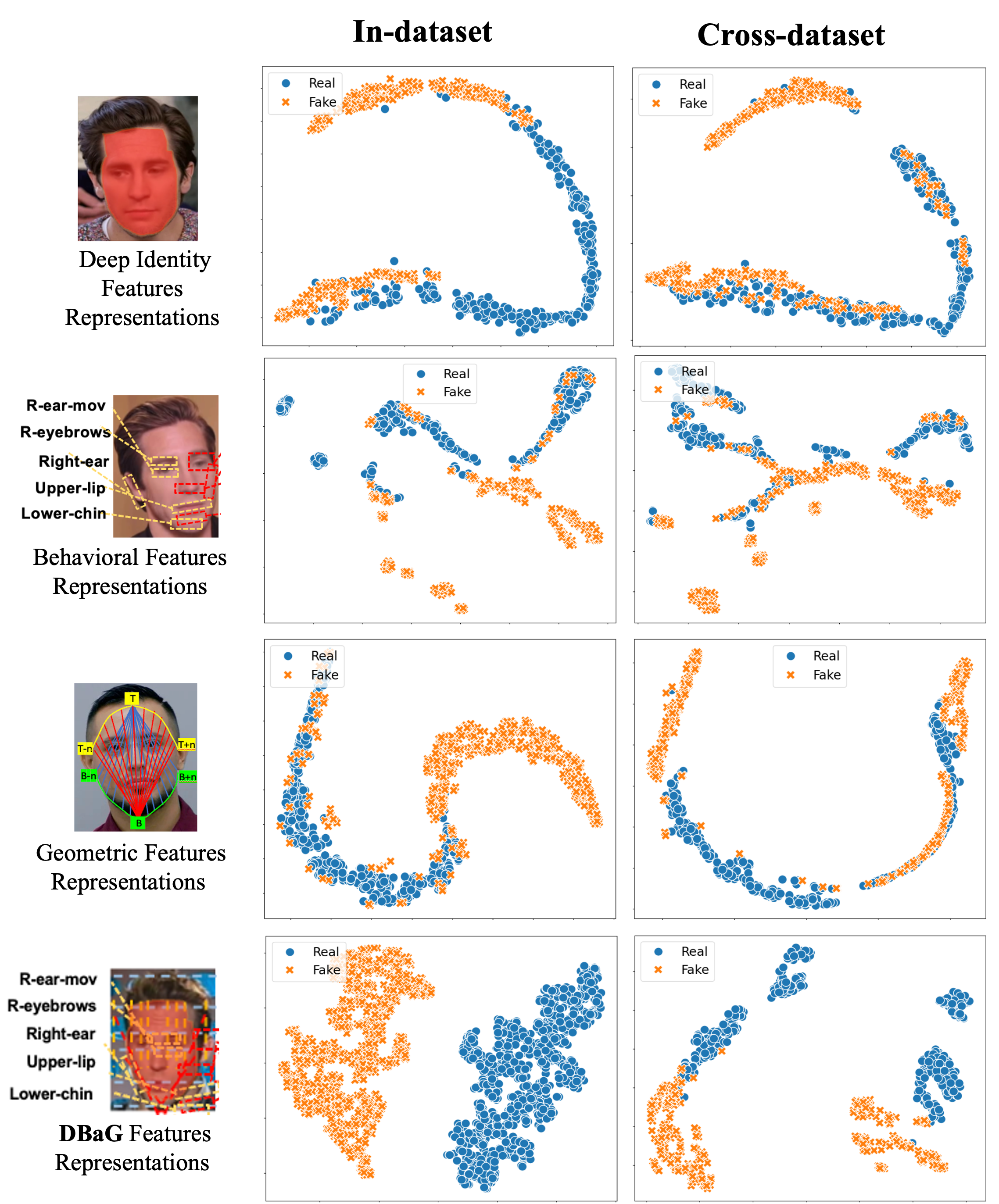}	
	\caption{
Analysis of in- and cross-dataset representations of individual and fused features. The left column highlights the facial regions where deep identity, behavioral, and geometric features are extracted. The scatter plots display the t-SNE visualization of real and fake samples using different feature types: deep identity in the top row, behavioral in the second, geometric in the third, and the fused DBaG (Deep identity, Behavioral, and Geometric) features in the bottom row. The DBaG fusion demonstrates superior discrimination between real and fake samples, enhancing classification performance and generalization across both within-dataset and cross-dataset tests.
 } \vspace{-10pt}
	\label{fig_1}%
  \end{center}
\end{figure}
Multimedia manipulation stems from techniques such as replacing facial expressions without altering the actor's identity \cite{Garrido_2014} and real-time expression transfer \cite{10.1145/2816795.2818056}. Moreover, the technology has advanced to the point where creating a deepfake now takes only 35 seconds, making the manipulation of multimedia faster and more accessible than ever before. Similarly, the year 2023 alone surge in the development of deepfake generation tools, with over 60\% of them being newly introduced within that year\footnote{\url{https://humanorai.io/deepfake-tools-statistics}}. The advancement in generative AI techniques, such as generative adversarial networks (GANs) and diffusion-based models, enabled users to produce highly realistic deepfake videos that are often indistinguishable from genuine videos.

Despite their benefits in fields such as entertainment and the film industry, deepfakes pose significant risks when used maliciously, threatening social integrity. The manipulation of faces in videos is particularly concerning, as seen in recent real-world incidents. For instance, a fabricated video of former President Obama mocking Donald Trump spread distrust, and a forged video of Ukraine's president urging soldiers to surrender sowed confusion and misinformation~\cite{xiao2022securing}. In the financial sector, deepfakes have also been weaponized, with a notable case involving a fraudulent transaction of \$25 million, triggered by a deepfake video of a chief financial officer during a video conference\footnote{\url{https://www.cnn.com/2024/02/04/asia/deepfake-cfo-scam-hong-kong-intl-hnk/index.html}}. For public figures, such fake content can lead to severe reputational damage and destabilizing consequences, amplifying the urgency to address the misuse of deepfake technology.


Although substantial progress has been made in the detection of deepfakes, there are still some key challenges that continue to limit the effectiveness of existing methods. The existing methods developed to encounter deepfake can be categorized into traditional solutions using handcrafted features and recent studies using deep learning models. These solutions have proven to be effective against specified types of deepfake but often fall short in terms of generalization. For example, Zhang et al.~\cite{zhang2017automated} employed Speeded Up Robust Features SURF descriptors for detecting face swap deepfakes, but their method is solely developed to detect deepfake images and ineffective when confronted with videos. In another study, Ciftci et al.~\cite{cciftcci2024deepfake} introduced biological signal analysis (e.g., heart rate) for forensic changes.
The approach presented in this study may suffer from limited generalization due to its reliance on a dataset that may introduce biases towards known generative models, specifically against unseen fake instances. Similarly, Jung et al.~\cite{jung2020deepvision} used eye-blinking anomalies for detection, but their method fails when subjects exhibit abnormal blinking patterns. These approaches, while valuable, demonstrate that single-feature-based solutions are limited in scope. On the other hand, traditional solutions such as those focusing on eye-blinking~\cite{li2018ictu} or behavioral biometrics~\cite{agarwal2020detecting} perform well in controlled scenarios but lack generalization to other types of forgeries or unseen datasets.

In contrast, recent deep learning solutions, such as those proposed by Li et al.~\cite{li2018exposing} and Agarwal et al.~\cite{agarwal2020detecting}, utilize facial landmarks and behavioral biometrics for deepfake detection. However, these models frequently encounter challenges when addressing diverse deepfake techniques, including lip-sync manipulations and unseen datasets. In another study, Fernandes et al.~\cite{fernandes2019predicting} employ biological signals by measuring heart rates through methods like skin color variation and Eulerian video magnification, which are subsequently analyzed using a Neural Ordinary Differential Equations (Neural-ODE) model. While this approach has shown effectiveness in detecting deepfakes, it is affected by increased computational complexity, potentially limiting its applicability in real-time scenarios. Similarly, Yang et al.~\cite{yang2021msta} introduce a multi-scale texture difference network that employed a ResNet-18 architecture to capture texture variations in manipulated images, although the presented solution achieve good performance on certain deepfakes these models often struggle with generalization across varying deepfake techniques, particularly in more complex situations. Other works~\cite{sabir2019recurrent,rossler2019faceforensics++} have explored CNN-based methods for the detection of swapped faces, however, there remains a critical need for more generalized solutions capable of consistently performing across diverse datasets and manipulation types. Collectively, these studies highlight the challenges faced by existing deep learning approaches in achieving comprehensive generalization,
revealing that reliance on isolated handcrafted features or end-to-end deep models alone is insufficient for effective detection.


To address the limitations of existing studies, we introduce a novel framework for video deepfake detection. The proposed framework leverages a unique combination of AR/VR-inspired behavioral features, golden ratio-based geometric features, and deep identity features to train the classification model. By employing a triplet input structure and a loss function that measures similarity, the presented framework assesses the closeness of each sample to either the real or deepfake class. Unlike existing methods that rely on single feature sets or binary classification, our framework integrates complementary DBaG features and trains using a similarity matrix, enhancing its ability to capture the complex characteristics of both in-dataset and cross-dataset deepfakes, as illustrated in Figure~\ref{fig_1}. For this analysis, we trained our system on FF++ for in-dataset testing and used CelebDF for cross-dataset analysis, visualized through t-SNE. To the best of our knowledge, this is the first work to utilize a comprehensive feature set, combining expert insights from AR/VR, golden ratio principles, and facial recognition for deepfake detection. The contributions of this work are summarized as follows:

\begin{enumerate}
    \item We introduce a novel descriptor named DBaG that encompasses deep identity, behavioral, and geometric information for deepfake detection. DBaG captures holistic aspects of a given video, leading to effective in- and cross-datasets deepfake detection.

    \item We present a novel video deepfake detection framework that employed a triplet-based classifier DBaGNet that effectively discriminates between authentic and manipulated content by leveraging distance-based learning on the comprehensive DBaG features. The triplet learning of DBaGNet helps in enhancing generalization, particularly for unseen deepfake examples, that improves the robustness of the proposed solution in cross-dataset evaluations.

    \item We performed rigorous experimentation to test the generalization ability of the presented framework on six different datasets: {FF++, DFD, DFDC, WLDR, CelebDF, and NVFAIR}. The results demonstrate that the combination of features in the DBaG descriptor is effective against vast types of deepfake in the seen and unseen cross-dataset settings.
 
    \item The comprehensive ablation study reveals that AR/VR-inspired behavioral features, golden ratio-based facial geometrical features, and deep identity features from facial recognition perform effectively within individual datasets. Moreover, the fusion of these behavioral, geometric, and deep identity features enhances the model generalization ability towards cross datasets evaluations.

\end{enumerate}

The rest of the paper is organized as follows: Section II reviews the literature on deepfake detection. Section III presents the details of the proposed framework. Section IV provides the evaluation details, including datasets and results. Lastly, Section V presents the conclusion and limitations of the paper.
\section{Literature Review}
The rapid advancement of deepfake generation methods raises concerns about the need for effective and generalized deepfake detection methods. The existing deepfake detection approaches can be broadly categorized into handcrafted and deep features based methods. Methods that employ handcrafted features mostly analyze inconsistencies in lip-syncing and blinking (behavioral) or inconsistencies in facial landmark movements (geometric). In contrast, deep learning techniques like CNNs excel at extracting subtle inconsistencies in facial features. When applied to video data, these methods utilize temporal information to identify inconsistencies across sequential frames. 
In the next subsections below we provide a brief overview of each category and discuss the recent studies conducted on enhancing generalization ability in deepfake detection.

\subsection{Traditional ML based deepfake detection}
Traditional deepfake detection methods mainly relied on handcrafted features, specifically designed to exploit signature vulnerabilities introduced by the generative algorithms. These features targeted inconsistencies induced by the deepfake generation algorithms, for instance, the authors of the study~\cite{yang2019exposing} revealed that deep fake videos often lack natural physiological signals like regular eye blinking and consistent head movements, which could serve as indicators for deep fake detection. In another study~\cite{matern2019exploiting}, the authors identified visual anomalies, such as changes in eye color, unconvincing reflections, and discrepancies in eye and tooth details, and used them to identify deepfakes. Similarly, the method proposed in~\cite{agarwal2019protecting} exploited distinct facial expressions. In~\cite{mccloskey2019detecting}, the method capitalized on generator limitations, while~\cite{qi2020deeprhythm} focused on subtle variations induced by the heartbeat on the face to discern video authenticity. Though these techniques were effective in their time, their dependence on specific cues
made them vulnerable to rapid advancements in deepfake technology. In contrast, some other works concentrate on methods for creating interpretable deep models through the use of self-explanatory mechanisms \cite{10.5555/3327757.3327875}. Instead of relying solely on raw inputs, these approaches used "units of explanation," such as abstract notions, to convey interpretability. This strategy was used for basic concept acquisition by Alvarez et al. in \cite{10.5555/3327757.3327875}, which was further used for case-based reasoning and prototype learning by Kim et al. \cite{10.5555/2969033.2969045}. Recently, the author of  \cite{10.5555/3454287.3455088} proposed ProtoPNet, which uses prototype learning to create predictions based on resemblance to class-specific image patches. However, although significant progress has been made in the development of intricate video classification models for explainable solutions (such as 3D CNNs, TSNs, and TRNs \cite{lee2023d,potok2023modulation}, there is a lack of interpretability in complex deepfake detection methods encompassing recent deepfake forgeries. 

\subsection{Deep learning based deepfake detection}
Deep learning approaches for deepfake detection offer more robust and adaptable solutions. Unlike traditional handcrafted features, neural networks are able to learn the complex patterns and subtle inconsistencies in real and fake faces, surpassing the limitations of manually engineered features. These solutions can be broadly categorized into frame-based, temporal, and spatial-temporal approaches. In frame-based deepfake detection, ~\cite{raza2023holisticdfd} proposed a deepfake detector that identified blending imperfections for video distortion detection. The author of the study~\cite{zhao2020detecting} presented a CNN-based model for detecting deformation artifacts. Similarly, the presented framework in~\cite{bibi2021digital} analyzed video segments to capture visual inconsistencies. In contrast, in temporal analysis,~\cite{nguyen2019multi} combined CNNs and LSTMs to analyze blinking patterns and temporal irregularities. The author of \cite{saealal2022using} used a spatio-temporal network for deepfake manipulated artifact detection. On the other hand, some studies developed lightweight deepfake detector models that can be used in computational intensive devices. For instance, the author of the study~\cite{tariq2019gan} presents ShallowNet based deepfake detector which excelled at detecting GAN-generated images even at low resolution. In another study~\cite{pellicer2024pudd}, Pellcier et.al., introduces a unified framework based on prototype learning for Deepfake Detection (PUDD), which uses similarity-based detection to identify deepfakes by comparing input data to known prototypes. The presented PUDD framework outperforms the existing methods by achieving 95.1\% accuracy on the Celeb-DF dataset. However, the presented solution was not tested against generalization across diverse datasets. Additionally, the model's effectiveness could be affected by sensitivity to prototype selection. Similarly, the study~\cite{agarwal2024deepfake} investigates the potential of deep neural network fusion for effective deepfake detection. The author emphasize the structural ability of various neural networks to capture distinct features for deepfake detection. For this, the author presents a multi-branch and multi-level architecture that separates fixed and adaptive knowledge from pre-trained networks which enhance the detection on low-power devices such as mobile devices. However, the generalization ability of this solution remains untested. In the next subsection, we explained and discussed the research conducted towards generalization of the deepfake detection.


\subsection{Deepfake detection toward generalization}
Generalization ability remains a critical hurdle in deepfake detection, affecting the performance of the deepfake detectors from adapting to unseen data. Despite this significant challenge, research specifically addressing generalization ability within this domain remains limited. Early attempts like FWA \cite{Li2018ExposingDV} exploited resolution differences between forged faces and backgrounds for detection. Recent studies, however, demonstrate significant progress towards enhanced generalizability. Face X-ray \cite{Li2019FaceXF} focuses on identifying blending boundary artifacts, while SPSL \cite{Liu2021SpatialPhaseSL} utilizes phase spectrum analysis in a frequency-based approach. Lip-Forensics \cite{Haliassos2020LipsDL} uses spatio-temporal networks to detect unnatural mouth movements, and SRM \cite{Luo2021GeneralizingFF} analyzes high-frequency noise for broader detection capabilities. 
In recent study~\cite{yan2024transcending}, The author Yan et.al., tackles the generalization issue in deepfake detection, where performance suffers from mismatched training and testing data distributions. The authors present the Latent Space Data Augmentation (LSDA) detector, which improves decision boundaries by simulating variations of forgery features in the latent space. The presented LSDA outperformed the existing detectors across multiple benchmarks. However, LSDA's reliance on effective simulation techniques may restrict its applicability to real-world scenarios with novel unseen deepfake forgeries. In another study~\cite{lai2024gm}, the authors propose the Generalized Multi-Scenario Deepfake Detection framework (GM-DF), which trains models on multiple datasets and utilizes a hybrid expert modeling approach along with a domain-aware meta-learning strategy. Although the framework shows promise in enhancing generalization, it may experience rapid declines in accuracy due to dataset discrepancies. Although these studies enhance the generalization of deepfake detection methods, their reliance on predetermined forgery patterns and consideration of the entire feature space makes them susceptible to disruption from irrelevant factors like background and identity.

Therefore, collectively both handcrafted and deep learning approaches offer distinct advantages: dataset oriented deepfake detection for handcrafted features and generalization ability for deep neural network. However, their fusion within a single framework holds significant promise for achieving both in-dataset and generalized deepfake detection. This strategy has been explored in few studies, some of which are discussed in the next subsections, with the aim of enhancing the generalization ability for deepfake detection systems.

\section{Proposed Framework}

In this section, we introduce the proposed framework for detecting video deepfakes. The presented framework is composed of three stages. The first stage involves preprocessing the video data to crop faces and extract facial landmarks. In the second stage, the extracted cropped faces and landmarks are passed through a multidisciplinary feature extraction process to obtain deep identity, behavioral, and geometric features. The obtained features are then fused together to create a comprehensive set of DBaG features. In the final stage, the fused DBaG features are input into a classifier, which uses a triplet loss objective to learn the representation of input video segments to classify them as real or fake based on a similarity index. The feature extraction process and the architecture of the proposed DBaGNet are presented in Figures~\ref{fig_3} and ~\ref{fig_4}. In the following subsections, we explain the process of each stage in detail. 


\subsection{Preprocessing}
Given a video $V$ consisting of $N$ frames, reshaped into $K$ batches ${S}^{B \times C \times W \times H}_{K}$, where $B$, $C$, $W$ and $H$ represent batch size, frame channels, width and height, respectively. A Multi-task Cascaded Convolutional Network ~\cite{zhang2016joint} is then applied on each batch $S^{B \times C \times W \times H}_{k},\text{ for } k = 1, 2, 3, \ldots, K-1$ to get the coordinates of faces in the input frames. Batches are reshaped into a sequence of frames ${V}^{N \times C \times W \times H}$ along with the coordinates of the detected faces. To ensure the quality of resulting cropped faces for the subsequent steps, we set  $120\times120$ as the acceptable facial dimensions required by the MTCNN, in the input video frames. Similarly, in the video with multiple faces, the face with larger dimensions is selected. 
After the face detection, faces are cropped and rescaled to $224\times224$ to get ${V}^{N \times C \times W \times H}_{224\times224}$, which is used for behavioral, geometric and deep identity features extraction.

\begin{figure*}[t]
    \centering
	\includegraphics[width=\textwidth]{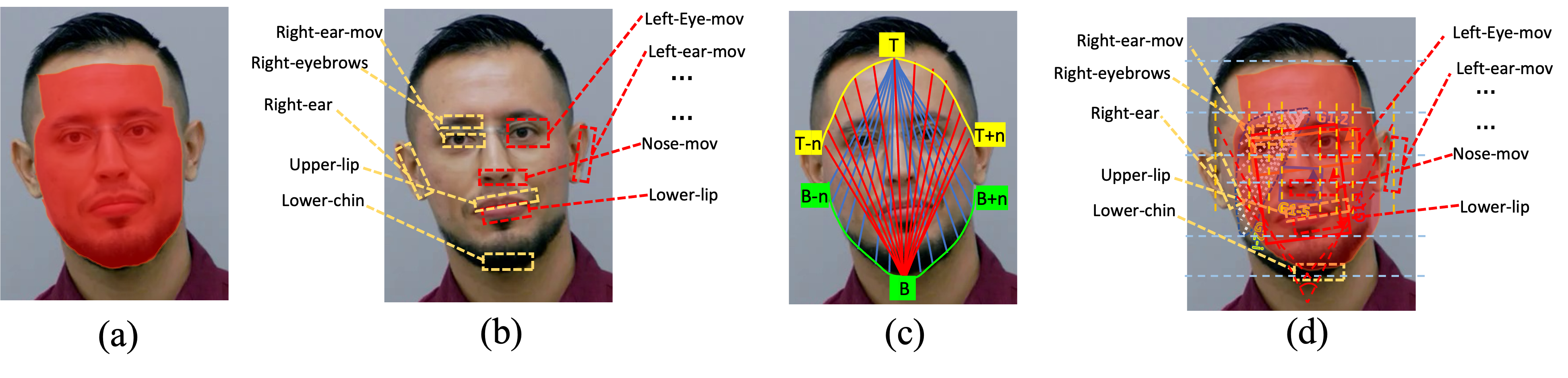}	
	\caption{
        Visual representation of the proposed feature descriptor, combining (a) deep identity features, (b) behavioral features, and (c) face geometry features to form the comprehensive DBaG descriptor (d).
  } \vspace{-10pt}
	\label{fig_2}%
\end{figure*}
\vspace{-10pt}

\subsection{ DBaG: A Multidisciplinary Feature Descriptor}
\begin{figure*}[t]
    \centering
    \includegraphics[width=14cm]{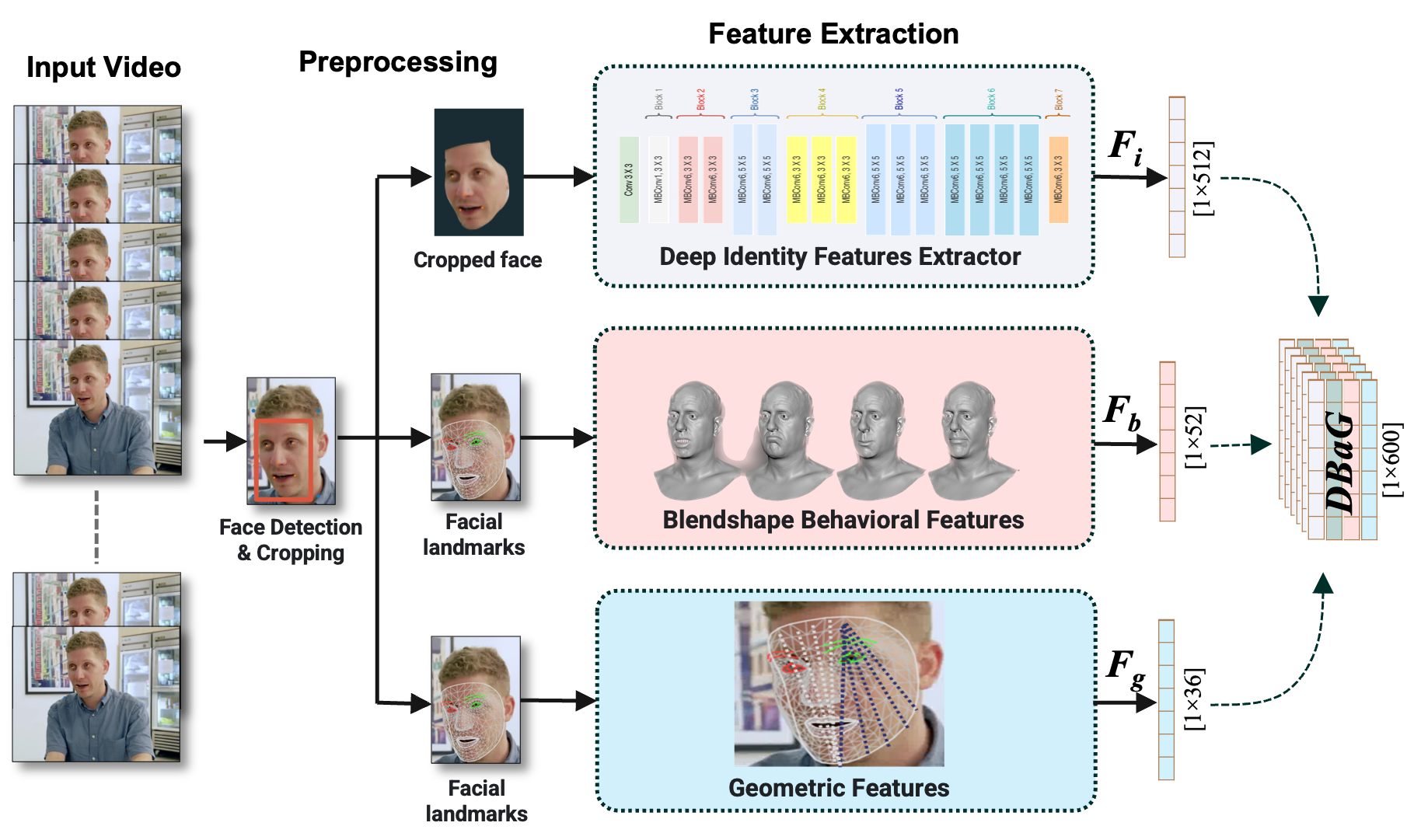}
    \caption{Detailed overview of the proposed feature descriptor. In preprocessing step, face detection and cropping are performed followed by feature extraction step where DBaG descriptor composed of Deep Identity, Behavioral and Geometric information is extracted.}
    \label{fig_3}
\end{figure*}

We argue that a holistic analysis of a face in a given video is more robust and generalized approach than a single cue based analysis. Based on our hypothesis and reported experimental results, we proposed a multidisciplinary DBaG descriptor composed of behavioral, facial geometry, and identity signature for deepfake detection. The visualization of DBaG features descriptor is given in Figure~\ref{fig_2} and the composition of the BDaG is visualized in Figure~\ref{fig_3}. The DBaG descriptor holistically analyze a given face for inconsistencies in facial behavior and face geometry along with relative identity information. Furthermore, the behavior and geometric features in DBaG descriptor are handcrafted, resulting in interpretability, while the identity features are extracted using deep models to detect low-level details like texture and skin color tone, and hence improve generalizability.  A detailed discussion on each component of the DBaG descriptor is given in the subsequent sections.

\subsubsection{Blendshape Features for Behavioral Analysis}
Facial behavioral patterns or expressions are the most complicated phenomena to mimic perfectly in deepfake generation. Inspired by the role of blendshape features that are commonly employed in virtual and augmented reality applications~\cite{menzel2022automated}, we use them to capture a wider spectrum of facial expressions for behavioral analysis. 
The acquisition of blendshape features starts with the extraction of facial landmarks. An input face frame $V_{224\times224}$ from ${V}^{N \times C \times W \times H}_{224\times224}$ is passed to a MobileNetV2-like customized architecture~\cite{MediaPipe_landmarker} ($M_{Net}$) to extract $478$ facial landmarks. Next, to capture facial expressions, a comprehensive set of 52-dimensional blendshape behavioral features is extracted with a MLP-Mixer backbone~\cite{MediaPipe_landmarker} ($MLP_{Mixer}$). A visual representation of behavioral feature calculation is given in Figure \ref{fig_2}(b). This approach facilitates the precise characterization of behavioral distinctions exhibited by the face throughout the frame sequence. Equations 1 \& 2 formulate the calculation of blendshape features as:


\begin{equation}
    LM_{[lm_1, lm_2, \ldots, lm_{478}]} = \sum_{j=1}^{N} M_{Net}[V_{224\times224}^{(j)}]
\label{eq1}
\end{equation}

\begin{equation}
    F_{b_{[1 \times 52]}} = \sum_{k=1}^{N} MLP_{Mixer}[LM_{[lm_1, lm_2, \ldots, lm_{478}^{(k)}]}]
\label{eq2}
\end{equation}

\subsubsection{Golden Ratio for Facial Geometry Analysis}
In some of the deepfake categories such as faceswap, the inner facial regions of a target face are replaced while the outer facial geometry is kept intact. To exploit the difference in inner and outer facial geometry, DBaG derives novel facial geometric features inspired by facial golden ratio ~\cite{meisner2018golden}. The facial golden ratio is originally used for beauty calculation and mostly covers the symmetry of a given face. To capture facial geometry information, we compute geometric features using facial landmarks to analyze temporal inconsistencies in facial structure. These inconsistencies may include subtle deviations in symmetry or proportion that arise when source and target facial geometries blend, creating a mix that appears visually appealing but lacks the natural cohesion of real faces in video frames.
For this task, we utilize Mediapipe~\cite{MediaPipe_landmarker} to capture facial geometry, by calculating distances formed by various landmarks. This facial geometry provides a holistic representation of the structural aspects when discerning genuine facial attributes. Specifically, we aim to capture alterations in facial geometry exhibited by an individual while speaking in a given video. Figure \ref{fig_2}(c) represents a visual representation of the derived geometric features. 

In our approach, we calculate the geometric features between specific sets of facial landmarks. Here, \( P = \{ p_1, p_2, \ldots, p_l \} \) represent the set of landmarks corresponding to the lower half of the outer face region, and \( Q = \{ q_1, q_2, \ldots, q_m \} \) represent the set of landmarks corresponding to the upper half of the outer face region. We define bottom of the face \( B \), as the central point of \( P \) and the top of face \( T \) as the central point of \( Q \). We compute distances from these central points to each landmark in the opposite region, covering a symmetric range of indices from \( T-n \) to \( T+n \) and \( B-n \) to \( B+n \) respectively:

\begin{equation}
    G_{T_{[1 \times 18]}} = \sum_{t=T-n}^{T+n} \left( |x_{B} - x_{t}| + |y_{B} - y_{t}| + |z_{B} - z_{t}| \right)
\end{equation}

\begin{equation}
    G_{B_{[1 \times 18]}} = \sum_{b=B-n}^{B+n} \left( |x_{T} - x_{b}| + |y_{T} - y_{b}| + |z_{T} - z_{b}| \right)
\end{equation}

These equations define the geometric features as distances from the central points \( B \) and \( T \) to the landmarks in the opposite region, spanning from \( T-n \) to \( T+n \) for $G_T$, and \( B-n \) to \( B+n \) for $G_B$.

The final feature vector representing the geometric properties is constructed as:

\begin{equation}
    F_{g_{[1 \times 36]}} = \left[ G_{T_{[1 \times 18]}}, \, G_{B_{[1 \times 18]}} \right]
\end{equation}

This approach provides a concise and focused calculation of distances between central points and surrounding landmarks in opposite facial regions, capturing geometric relationships precisely.


\subsubsection{Deep Identity Features Signature Analysis }
In addition to behavioral and geometric features ,  DBaG also include deep identity facial features to capture individual's facial characteristics or signature. In order to capture deep identity signature, we deploy a quality adaptive face-recognition model, AdaFace~\cite{kim2022adaface}, which uses a ResNet backbone and applies a margin based loss function to capture the deep features in low quality images as given in Equation 6.

\begin{equation} 
\theta = -\log\left(\frac{\exp(g(\Theta_{yi}, m))}{\exp(g(\Theta_{yi}, m)) + \sum_{j\neq yi}^{n} \exp(s \cos\Theta_j)}\right)
\label{eq6}
\end{equation}
	
where $\theta j$ denotes the angle between the feature vector of current frame and the $j^{th}$ classifier weight vector, $yi$ represents the index of the real face label, and $m$ corresponds to the margin, serving as a scalar hyperparameter. Finally, a feature vector of size 512 is obtained as given in Equation 7. 

\begin{equation}
     F_{i_{[1 \times 512]}}  = \sum_{m=1}^{N} M_{AF}(\theta ([V_{224\times224}])) 
\label{eq7}
\end{equation}
			
where $V_{224\times224}$ is the input image, $\theta$ is the loss function as shown in \ref{eq6}, and $M_{AF}$ is the transformation function used in AdaFace to compute the final feature vector.




Our DBaG feature descriptor brings different aspects, including behavioral $F_b$ showing how the face expresses itself, geometry $F_g$, revealing the face's structure and identity signature $F_i$, uncovering hidden patterns from generative algorithms.
Jointly, these features form our final feature vector, providing a well-rounded view of the information needed to tell if a video is real or fake. Equation~\ref{eq9} formulates the final feature vector of DBaG.
\begin{equation}
Fv_{ [1 \times 600]} = [F_b+F_g+F_i]
\label{eq9}
\end{equation}
To capture the spatial and temporal details in the video frames, we reshape the DBaG feature descriptor vectors to 2D slices of 120 frames with an overlap of 60 frames. The final input vector for the model is $Fv_{[120 \times 600]}$.

\subsection{Classification}
\begin{figure*}[t]
    \centering
    \includegraphics[width=\textwidth]{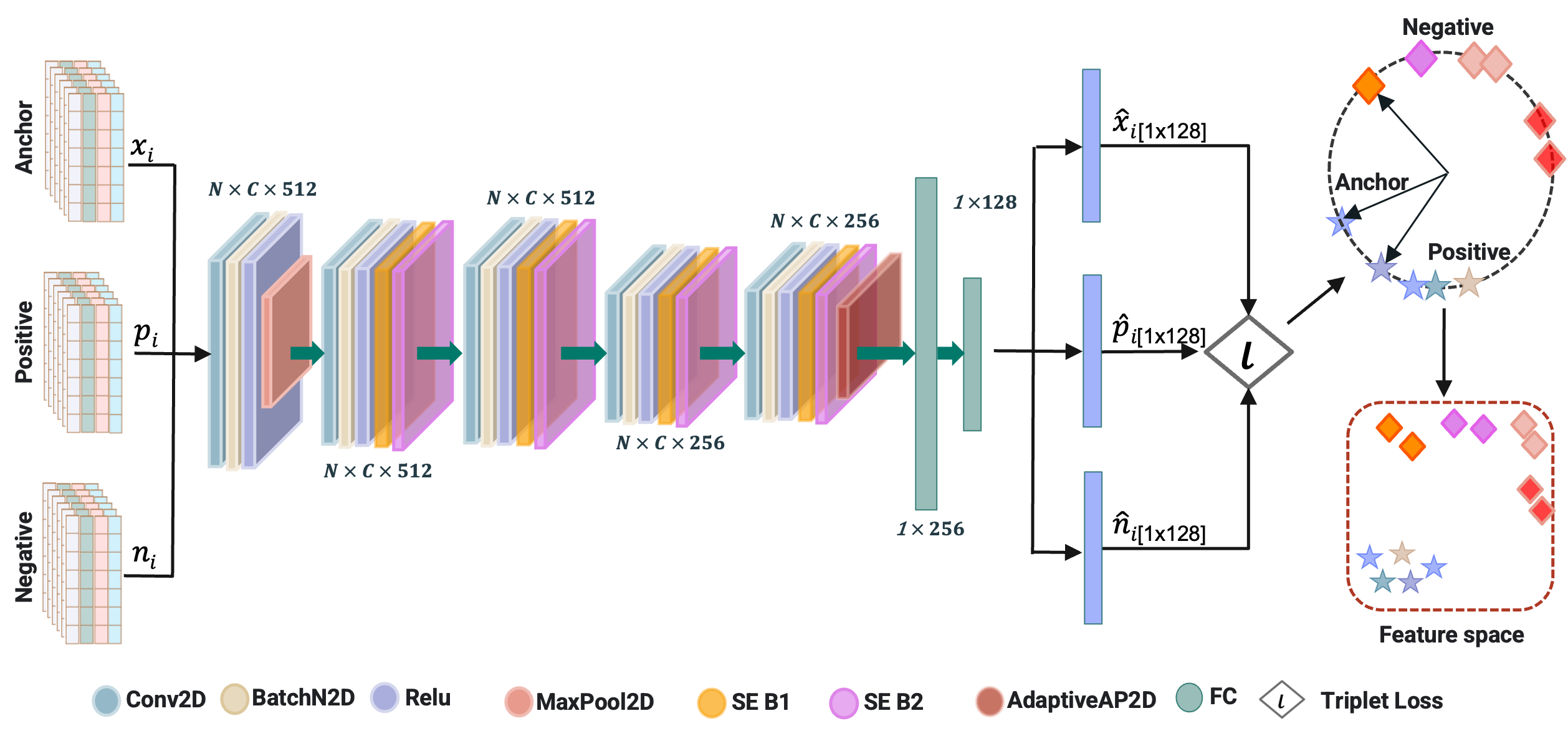}
    \caption{Detailed architecture of the proposed DBaGNet with triplet loss for representation learning.}\vspace{-15pt}
    \label{fig_4}
\end{figure*}

To perform classification in an efficient manner, we designed a deep learning classifier DBagNet as shown in Figure~\ref{fig_4}, which employs residual learning alongside both spatial and temporal domains. This architecture incorporates attention-based mechanisms and pooling techniques, drawing on prior research \cite{he2016deep, hu2018squeeze}  to create a model that can effectively generalize across manipulated media.
\subsubsection{Representation Learning with triplet loss}
Prior to representation learning, the obtained DBaG features passed to the DBaGNet consists of five squeeze and excitation based residual layers. 
The DBaGNet initial block consists of a 2D convolution with a 7×7 kernel and a stride of 2, followed by batch normalization, ReLU activation, and a max pooling layer with a 3×3 kernel. This block downscales the input and extracts low-level spatial features. Representational learning is primarily carried out through residual blocks, each of which contains two 3×3 convolutions with batch normalization and ReLU activation. Each block also integrates a squeeze-and-excitation (SE) component for adaptive feature recalibration, applied after the second convolution.
The final convolution layers are followed by an adaptive average pooling layer, which standardizes the feature map size (1×1), regardless of the original input size. After the convolutional and pooling layers, two fully connected layers reduce the dimensionality and generate an embedding vector. These layers use Leaky ReLU activations to maintain feature sensitivity and ensure stable gradient flow, especially important for effective embedding vectors. 

Following this, the proposed DBaGNet is trained with triplet margin loss to learn the triplet embeddings for generalized deepfakes detection. The training with triplet learning objective, requires the dataset to be constructed to have a triplet based input having an anchor $x_i$, a positive $p_i$ and a negative $n_i$ features. Next, the DBaG features for anchor, positive and negative samples are transformed to $\hat{x}_i$, $\hat{p}_i$ and $\hat{n}_i$ by feeding them into residual dense layer architecture to obtain finalized feature vectors. The resultant $\hat{p}_i$ has the same label as the $\hat{x}_i$ in contrary the $\hat{n}_i$ has different label. Next, we used triplet margin loss as the objective function to generate the discriminative embeddings of real and fake feature vectors. The triplet margin loss helps avoid over-fitting to the training data and better generalization. The triplet margin loss can be defined as:
\begin{equation}
    \mathcal{L_{\text{triplet}}}(\hat{x}_i,\hat{p}_i,\hat{n}_i) = \max\{D(\hat{x}_i, \hat{p}_i) - D(\hat{x}_i, \hat{n}_i) + \text{m}, 0\}
\end{equation}
where $\hat{x}_i,\hat{p}_i$ and $\hat{n}_i$ are anchor, positive and negative representation vectors. $m$ is the margin hyperparameter.  The margin ensures that the distance between the anchor and the negative vector is not just greater than the distance between the anchor and the positive vector, but it is greater by at least the specified margin. A larger margin enforces more significant separation of positive and negative vectors, which can lead to better discrimination but might make training more challenging. $D$ is the pairwise euclidean norm between anchor, positive and anchor, negative pairs, and is defined as:
\begin{equation}
    D(\hat{x}_i, \hat{p}_i) = \|\hat{x}_i - \hat{p}_i\|_2 = \sqrt{\sum_{i} (\hat{x}_i - \hat{p}_i)^2}
\end{equation}
\begin{equation}
    D(\hat{x}_i, \hat{n}_i) = \|\hat{x}_i - \hat{n}_i\|_2 = \sqrt{\sum_{i} (\hat{x}_i - \hat{n}_i)^2}
\end{equation}

During the training phase, model learns the representations of triplets in the form of anchor, positive and negative class. After the training process is complete, embedding vectors of the training samples are saved as the reference representations along with their labels. 

\subsection{Testing}
To evaluate the model, the testing samples are passed through the trained network, generating embeddings for each slice. We implement a similarity comparison approach from the reference set for the labels prediction of test embeddings. We calculate the distance of each test embedding from the embeddings of reference set to create a distance matrix.  The final label of test embedding is predicted based on the majority labels of m closest neighbors in the reference set. 

Once the training process is complete, we generate embeddings for each sample in the training set, referred to as the reference set $E_{\text{ref}}$. $E_{\text{ref}}$ is stored as embeddings, labels pairs $ \{ (e_1,y_1), (e_2,y_2) \ldots, (e_n,y_n) \}$, for n reference samples. These embeddings form a basis against which new (unseen) test samples are compared, providing a fixed point of reference for label prediction.
For each sample in the test set, the model computes a new embedding  $e_{test}$ by passing it through the trained DBagNet. This embedding represents the learned features of the test sample in the same embedding space as the reference set, capturing characteristics that differentiate real from fake data.To determine the label of $e_{test}$, we calculate its Euclidean distance from each reference embedding $e_i$ in $E_{ref}$, providing a measure of similarity. The Euclidean distance $d_i$ between $e_{test}$ and $e_i$
 is given by:
\begin{equation}
d_i = \| e_{\text{test}} - e_i \|_2 = \sqrt{\sum_{j} (e_{\text{test}, j} - e_{i, j})^2}    
\end{equation}
where j indexes the dimensions of the embedding vectors. This step produces a set of distances ${d_1,d_2,d_3,...,d_n}$ that indicate how close the test embedding is to each reference sample. To assign a label to $e_{test}$, we rank the distances ${d_1,d_2,d_3,...,d_n}$ in ascending order and identify the $m$ smallest distances, representing the nearest neighbors of $e_{test}$ in the embedding space. The label is then determined by a majority vote among the labels of these nearest neighbors. Formally, the predicted label  is given by:
\begin{equation}
y_{\text{pred}} = M_{vote}(\{y_{i_1}, y_{i_2}, \ldots, y_{i_m}\})
\end{equation}
where ${i_1}, y_{i_2}, \ldots, y_{i_k}$ are the indices of the $m$ nearest embeddings in $E_{ref}$ and $M_{vote}$ selects the label that appears most frequently among these neighbors. This majority voting process ensures that the final prediction takes into account multiple neighbors, providing robustness against outliers and minor variations in the embedding space.

\section{Experimentation and results}
This section discusses the details of the evaluation criteria, validation experiments on benchmark deepfake datasets including comparison with State-of-the-art methods, and the ablation study based on different combinations of feature sets and cross-validation experiments to evaluate the generalizability of the proposed framework.

\begin{table}
  \centering
  \caption{Performance evaluation of the proposed method on DFDC, WLDR, DFD, CelebDF, and face swapping subsets of FF++.}
  \begin{tabular}{l c c c}
    \hline
    \textbf{Dataset} & \textbf{ACC} & \textbf{AUC} & \textbf{EER} \\
    \hline
    DFDC & 97.71 & 96.78 & 0.72 \\
    \hline
    WLDR & 99.31 & 99.29 & 0.93 \\
    \hline
    DFD & 95.15 & 93.07 & 7.65 \\
    \hline
    CelebDF & 99.71 & 99.43 & 0.46 \\
    \hline
    FF-FaceSwap & 99.37 & 99.12 & 2.42 \\
    \hline
    FF-Deepfake & 99.18 & 99.02 & 1.02 \\
    \hline
    FF-FaceShifter & 99.49 & 99.23 & 1.23 \\
    \hline
    FF-Combined & 99.47 & 99.41 & 1.82 \\
    \hline
  \end{tabular}
  \label{Table1}
\end{table}

\subsection{Datasets}
We evaluate the feasibility and superiority of the proposed framework on six large-scale datasets of manipulated samples from recent GAN's and Diffusion models, confirming a well-suited evaluation to assessing the model's efficacy for generalization on new and old types of deepfakes. Short description of each dataset is given below.

\subsubsection{DFDC~\cite{dolhansky2020deepfake}} The DFDC is a large-scale dataset, with over 100k videos generated using 8 different GAN-based algorithms. We randomly selected 6800 videos with 1050 real and 5750 fake videos.

\subsubsection{CelebDF-v2~\cite{Celeb_DF_v2_cvpr20}} CelebDF-v2 comprising of 590 real videos and 5639 fake videos of  59 celebrities generated with recent generative models.

\subsubsection{World Leaders Dataset (WLDR)~\cite{agarwal2019protecting}} WLDR composed of five US presidential candidates including Bernie Sanders, Barack Obama, Elizabeth Warren, Joe Biden and Hillary Clinton. This dataset is comparatively challenging because the Deepfakes in WLDR are generated using impersonations of each political figure. This dataset contains 595 real and 82 fake videos.

\subsubsection{ DFD~\cite{dfd_2019}}
The deepfake detection dataset (DFD) by Google, in collaboration with Jigsaw, is generated with latest GAN algorithms. This dataset contains 363 high quality real and 3068 face-swap fake videos of 28 paid actors. This dataset has videos with one and more than one individual, performing different tasks, like talking, walking and hugging, and different emotional expressions like anger, disgust, happiness and neutral.

\subsubsection{FaceForensics$++$ (FF$++$) ~\cite{rossler2019FF}}
FF$++$ dataset stands as the most extensively employed forgery dataset. This dataset is comprised of five manipulation techniques, including three swapping techniques (DeepFakes, FaceSwap and FaceShifter), and two expression swapping techniques (Face2Face and NeuralTexture). This dataset contains $1000$ real videos and $1000$ for each fake technique with a total of $6000$ videos.

\subsubsection{NVIDIA Facial Reenactment (NVFAIR)~\cite{prashnani2023avatar}}
 NVFAIR is a new expression swapping dataset recently generated by NVIDIA. This dataset contains hours of expressions swapped videos of 161 identities, 24 from RAVDESS, 91 from CREMA-D, and 46 from their own video-conferencing data. They used three reenactment algorithms such as Face-vid2vid, LIA, and TPS to generate manipulated samples.

\subsection{Evaluation Metrics and Experimental Setup}
In our evaluation, we used Accuracy (ACC), equal error rate (EER) and area under the curve (AUC) metrics. A random split of 80\%:20\% is used for training and testing, respectively for DFDC, DFD, WLDR, and NVFAIR as these datasets have enough videos for each identity. While for the FF++ and CelebDF datasets we followed a different approach for train/test split, similar to~\cite{agarwal20behavior}, because these datasets have a limited number of videos per identity. Therefore, we chose the training and testing subsets from each video as follows: the first 20\% of the video segment is used for training, the last 20\% for testing, and the remaining video is ignored to prevent overlapping. 



\subsection{Performance Evaluation of the Proposed Framework}
To evaluate the performance of the proposed framework, we tested our framework on the two most common types of deepfake video i.e., face swapping and expressions swapping. A detailed discussion on the obtained results is presented in subsequent sections.

\subsubsection{Performance Analysis on Deepfakes}
This experiment is designed to analyze the performance of the proposed framework on the most recent and challenging deepfakes on the standard benchmarks. We used DFDC, CelebDF, WLDR, DFD, and face-swapping subsets of FF$++$ for this experiment. The results are presented in Table~\ref{Table1}. It can be seen that the proposed framework achieves remarkable performance with an AUC of 93\% to 99\%, on DFDC, WLDR, CelebDF and FF++ datasets. The performance is comparatively low on DFD dataset, due to the presence of more than one face in some videos. The proposed framework is robust to the new types of deepfakes and achieves remarkable performance on new types of manipulations like FaceShifter and DFDC datasets.

\begin{figure*}[t]
	\centering 
	\includegraphics[width=0.90\textwidth]{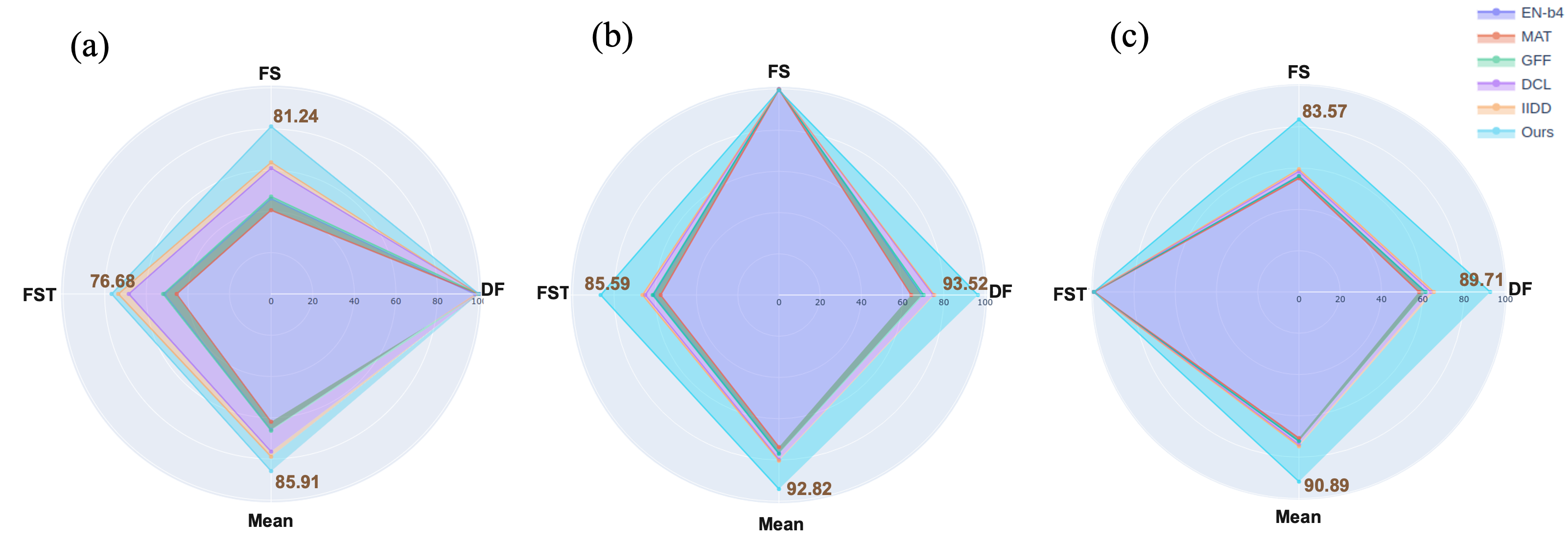}	
	\caption{Effectiveness of proposed framework on cross-manipulation evaluation: (a) shows the effectiveness of model when trained on DeepFake (DF) and tested on FaceSwap (FS) and Face Shifter (FST) subsets of FF++ and test part of FS. In (b) \& (c) model trained on FS and FST and tested on the other two subsets.}\vspace{-10pt}
	\label{fig_5}%
\end{figure*}

\subsubsection{Performance Analysis on Expressions Swaps}
To analyze the performance evaluation on expressions swaps, we conducted experiments on NVFAIR and expression swapping subsets of the FF$++$ dataset. The results in Table~\ref{Table2} demonstrate the performance of our framework on facial reenactment. The proposed framework achieves remarkable performance on facial reenactment, including the new types in the NVFIAR dataset, with an AUC of 96\% to 98\%. Performance is comparatively low on the NeuralTextures type of expression swaps. This indicates the detection of NeuralTextures expressions swap is a challenging task.

\begin{table}
\centering
\caption{Performance evaluation of the proposed method on expressions swapping subsets of FF++ and NVFAIR Dataset.}

\begin{tabular}{l c c c} 
 \hline
 Manipulation Type & ACC  & AUC & EER \\
 \hline
 NVFAIR-FaceV2V & 97.18 & 96.27 & 6.12 \\ 
 \hline
 NVFAIR-RAVDESS & 98.08 & 97.62 & 5.96 \\ 
 \hline

 FF-Face2Face &  98.82 & 98.24 & 3.17 \\ 
 \hline
 FF-NeuralTextures &  97.27 & 96.38 & 4.21 \\ 
 \hline
 FF-Combined &  98.28 & 97.86 & 3.53 \\ 
 \hline
\end{tabular}
\label{Table2}
\end{table}

\subsection{Comparative Performance Analysis with State-of-the-art}
The proposed framework is compared with current state-of-the-art (SOTA) techniques to analyse its effectiveness over various datasets. We compared the proposed framework with PWL~\cite{agarwal2019protecting} and AnB~\cite{agarwal20behavior} on WLDR dataset while 2-stream~\cite{zhou2017two}, XceptionNet~\cite{Celeb_DF_v2_cvpr20},Head Pose~\cite{yang2019exposing}, MesoNet~\cite{afchar2018mesonet}, DSP-FWA\cite{li2018exposing}, AnB~\cite{agarwal20behavior}, MAT~\cite{zhao2021multi}, SLADD~\cite{chen2022self}, Face-x-ray~\cite{li2020face}, PCL+12G~\cite{zhao2021learning}, and SBI~\cite{shiohara2022detecting} on the rest of the datasets. Table~\ref{Table3} gives a comparative performance analysis. As seen, the proposed framework outperform PWL and AnB on the WLDR dataset, achieving an AUC of 99.29. It also achieves an AUC of 99.41 on FF++, 96.78 on DFDC, and 99.43 on the CelebDF dataset, outperforming other methods on DFDC and CDF datasets. The proposed method demonstrated strong performance, achieving AUC scores of 99.41\% on the FF++ dataset and 92.12\% on the DFD dataset, positioning it as the second-highest performer and closely matching the leading results of 99.64\% and 93.2\% attained by SBI and AnB, respectively.

\begin{table}
  \centering
\caption{Comparison with the SOTA on multiple benchmarks. Performance metric used for comparison is AUC.
}
\begin{tabular}{c c c c c c} 
 \hline
 Methods                      & WLDR & FF++ & DFD & DFDC & CDF \\
 \hline
 PWL~\cite{agarwal2019protecting}                      & 	93.00 & -   & -     & -    & - \\ 
 \hline
 2-stream~\cite{zhou2017two}                & -   & 70.1 & 52.8  & 61.4 & 53.8 \\ 
 \hline
 XceptionNet~\cite{Celeb_DF_v2_cvpr20}             & -   & 99.7 & 85.9  & 72.2 & 65.3 \\ 
 \hline
 Head Pose~\cite{yang2019exposing}               & -   & 47.3 & 56.1  & 55.9 & 54.6 \\ 
 \hline
 MesoNet~\cite{afchar2018mesonet}                  & -   & 84.7 & 76.0  & 75.3 & 54.8 \\ 
 \hline
 DSP-FWA\cite{li2018exposing}                 & -   & 93.0 & 81.1  & 75.5 & 64.6 \\ 
 \hline
 AnB~\cite{agarwal20behavior}                     & 99.0 & 99.2 & \textbf{93.2}  & 95.6 & 99.3 \\ 
 \hline
 MAT~\cite{zhao2021multi}     & - & 99.61 & -  & - & - \\ 
 \hline
 SLADD~\cite{chen2022self}                     & - & 98.40 & -  & - & - \\ 
 \hline
 Face-x-ray~\cite{li2020face}               & - & 99.17    & -  & - & - \\ 
 \hline
 PCL+12G~\cite{zhao2021learning}                  & - & 99.11    & -  & - & - \\ 
 \hline
 SBI~\cite{shiohara2022detecting}                      & - & \textbf{99.64}     & -   & - & - \\ 
 \hline
 Ours                  & \textbf{99.29} & 99.41 & 92.12 & \textbf{96.78} & \textbf{99.43} \\ 
 \hline
\end{tabular}
\label{Table3}
\end{table}

\begin{table}
\centering
\caption{Cross-manipulation evaluation. DeepFakes, FaceSwap and FaceShifter manipulation sets of FF++ are denoted by DF, FS and FST, respectively.
}
\begin{tabular}{c c c c c c}
 \hline
 Train & Method & DF & FS & FST & Mean \\
 \hline
 DF  & EN-b4~\cite{tan2019efficientnet} & 97.97        & 46.24          & 51.26 & 65.82 \\ 
     & MAT~\cite{zhao2021multi}  & 99.92         & 40.61          & 45.39 & 61.97 \\ 
     & GFF~\cite{luo2021generalizing}  & 99.87         & 47.21         & 51.93 & 66.34 \\ 
     & DCL~\cite{sun2022dual}  & \textbf{99.98} & 61.01        & 68.45 & 76.48 \\ 
     & IID~\cite{huang2023implicit}  & 99.51        & 63.83         & 73.49 & 78.94 \\ 
     & Ours  & 99.82        & \textbf{81.24}  & \textbf{76.68} & \textbf{85.91} \\ 
 \hline
 FS  & EN-b4~\cite{tan2019efficientnet} & 69.25        & 99.89          & 60.76 & 76.63 \\ 
     & MAT~\cite{zhao2021multi}  & 64.13         & 99.67          & 57.37 & 73.72 \\ 
     & GFF~\cite{luo2021generalizing}  & 70.21         & 99.85          & 61.29 & 77.12 \\ 
     & DCL~\cite{sun2022dual}  & 74.80         & \textbf{99.90}  & 64.86 & 79.85 \\ 
     & IID~\cite{huang2023implicit}  & 75.39        & 99.73          & 66.18 & 80.43 \\ 
     & Ours  & \textbf{93.52} & 99.34        & \textbf{85.59} & \textbf{92.82} \\ 
 \hline
 FST  & EN-b4~\cite{tan2019efficientnet} & 61.11          & 56.19          & \textbf{99.52} & 72.27 \\ 
     & MAT~\cite{zhao2021multi}  & 58.15            & 55.03          & 99.16 & 70.78 \\ 
     & GFF~\cite{luo2021generalizing}  & 61.48            & 56.17          & 99.41 & 72.35 \\ 
     & DCL~\cite{sun2022dual}  & 63.98            & 58.43          & 99.49 & 73.97 \\ 
     & IID~\cite{huang2023implicit}  & 65.42           & 59.50          & 99.50 & 74.81 \\ 
     & Ours  & \textbf{89.71} & \textbf{83.57}  & 99.37 & \textbf{90.89} \\ 
 \hline
\end{tabular}

\label{Table5}
\end{table}

\begin{table*}
\centering
\caption{Performance analysis of Cross-dataset evaluation. Model trained on FF$++$(c23) and tested on CelebDF, DFD and DFDC.}
\begin{tabular}{l c c c}
\hline
Method & Celeb-DF & DFD & DFDC \\
 & AUC(\%) \hspace{3pt} EER(\%) & AUC(\%) \hspace{3pt} EER(\%) & AUC(\%) \hspace{3pt} EER(\%) \\
\hline
XceptionNet~\cite{Celeb_DF_v2_cvpr20} & 65.27 \hspace{5pt} 38.77 & 65.27 \hspace{5pt} 38.77 & 87.86 \hspace{5pt} 21.04 \\
\hline
EN-b4~\cite{tan2019efficientnet} & 68.52 \hspace{5pt} 35.61 & 68.52 \hspace{5pt} 35.61 & 87.37 \hspace{5pt} 21.99 \\
\hline
Face X-ray~\cite{li2020face} & 74.20 \hspace{6pt} - & 85.60 \hspace{6pt} - & 70 \hspace{6pt} - \\
\hline
MLDG~\cite{li2018learning} & 74.56 \hspace{5pt} 30.81 & 88.14 \hspace{5pt} 21.34 & 71.86 \hspace{5pt} 34.44 \\
\hline
F3-Net~\cite{wei2020f3net} & 71.21 \hspace{5pt} 34.03 & 86.10 \hspace{5pt} 26.17 & 72.88 \hspace{5pt} 33.38 \\
\hline
MAT~\cite{zhao2021multi} & 76.65 \hspace{5pt} 32.83 & 87.58 \hspace{5pt} 21.73 & 67.34 \hspace{5pt} 38.31 \\
\hline
GFF~\cite{luo2021generalizing} & 75.31 \hspace{5pt} 32.48 & 85.51 \hspace{5pt} 25.64 & 71.58 \hspace{5pt} 34.77 \\
\hline
LTW~\cite{sun2021domain} & 77.14 \hspace{5pt} 29.34 & 88.56 \hspace{5pt} 20.57 & 74.58 \hspace{5pt} 33.81 \\
\hline
Local-relation~\cite{chen2021local} & 78.26 \hspace{5pt} 29.67 & 89.24 \hspace{5pt} 20.32 & 76.53 \hspace{5pt} 32.41 \\
\hline
DCL~\cite{sun2022dual} & 82.30 \hspace{5pt} 26.53 & 91.66 \hspace{5pt} 16.63 & 76.71 \hspace{5pt} 31.97 \\
\hline
UIA-ViT~\cite{zhuang2022uia} & 82.41 \hspace{6pt} - & \textbf{94.68} \hspace{6pt} - & 75.80 \hspace{6pt} - \\
\hline
IID~\cite{huang2023implicit} & 83.80 \hspace{5pt} 24.85 & 93.92 \hspace{5pt} 14.01 & 81.23 \hspace{5pt} 26.80 \\
\hline
Ours & 82.54 \hspace{5pt} 52.24 & 82.33 \hspace{5pt} 23.12 & 85.94 \hspace{5pt} 18.00 \\
\hline
Ours + Aug  & \textbf{89.72 \hspace{5pt} 8.83} & 91.62 \hspace{5pt} \textbf{7.45} & \textbf{92.05 \hspace{5pt} 6.51} \\
\hline
\end{tabular}
\label{Table4}
\end{table*}

\subsection{Cross-manipulation evaluation}
To analyze the generalizability of the proposed framework on unseen type of deepfakes for the seen identities, we compared the performance against the SOTA on FaceShifter(FST), FaceSwap(FS) and DeepFakes(DF) manipulation classes of FF++. The model was trained on each of the three types of deepfakes (FST, FS, and DF) and tested on the test set of the same and the other two subsets. It can be observed in the Table~\ref{Table5} and Figure~\ref{fig_5} that the mean AUC achieved by our framework is significantly higher compared to the SOTA, with the mean AUC gains of 6.97, 12.39 and 16.08 respectively. The AUC gain for DF training is comparatively low due to the low performance on FST as shown in Figure~\ref{fig_5} $(c)$. The reason for this performance drop is the difference in quality and the nature of these two manipulation algorithms. Videos generated with DF do not have natural eye blinking and lip movements, while the videos generated with FST have more realistic behavioral characteristics, making them more difficult to detect.

\subsection{Cross-dataset evaluation}
To analyze the generalizability of the proposed framework, we performed a comparative analysis with the SOTA by training the model on FF++ and testing on CelebDF, DFD and DFDC datasets. To conduct this experiment, the proposed framework was trained on the FF++(C23) and tested on test sets of other three datasets. We selected SOTA methods for comparison, including XceptionNet~\cite{Celeb_DF_v2_cvpr20}, MLDG~\cite{li2018learning}, MAT~\cite{zhao2021multi}, DCL~\cite{sun2022dual}, EN-b4~\cite{tan2019efficientnet}, GFF~\cite{luo2021generalizing} and IID~\cite{huang2023implicit}. It is shown in Table~\ref{Table4}, the proposed model achieves superior performance over the SOTA.

\begin{table}
  \centering
\caption{Effectiveness of different combination of features on DFDC, WLDR, FF$++$, and NVFAIR. FF-FS refers to Face swapping manipulations and FF-ES refers to expressions swapping manipulations.
}
\begin{tabular}{l c c c c c} 
 \hline
 Features               & DFDC  &  WLDR  & FF-FS & FF-ES & NVFAIR     \\
 \hline
 DIF                    & 84.14 & 92.42  & 85.49 & 87.24  &  88.30   \\
 \hline
 DIF+B                  & 93.79 &  95.37 & 90.69 & 92.14 & 94.55     \\
 \hline
 DIF+B+G          & \textbf{96.78} & \textbf{99.29}  & \textbf{94.18}   & \textbf{97.18} & \textbf{99.21}       \\
 \hline
\end{tabular}

\label{Table6}
\end{table}

\subsection{Ablation study}
To analyze the effectiveness of extracted features, we performed an ablation study to evaluated our framework with the different combinations of feature sets on the DFDC, WLDR, FFF++ face swap, FF++ expression swap and NVFIAR datasets. The quantitative results on each dataset are presented in Table~\ref{Table6}.
To evaluate the effectiveness of different feature combinations on our model, we performed an ablation study across multiple datasets, analyzing the performance both within individual datasets and across datasets. Specifically, we examined our model on DFDC, WLDR, FF++ (face swap and expression swap), and NVFIAR datasets, testing each feature combination's impact. Results for in-dataset evaluations are presented in Table~\ref{Table6}, while cross-dataset results are provided in Table~\ref{Table7}. 

Starting with only deep identity features (DIF), the model established a baseline by achieving $88 \pm 5\%$ AUC. With the addition of behavioral features, along with deep identity features, our model improves 9.65\%, 2.95\%, 5.20\%, 4.90\% and 6.25\% AUC on DFDC, WLDR, FF-FS, FF-ES and NVFAIR, respectively. The AUC gains for geometric features, combined with deep identity and behavioral features, are 4.64\%, 3.87\%, 3.49\%, 5.04\% and 4.66\%. Table~\ref{Table6} clearly shows that the final feature set, with behavioral and geometric features, achieves a more stable performance on all datasets. For cross-dataset analysis, we trained the model on FF++ and evaluated it on DFDC, DFD, and CelebDF. For cross-dataset analysis, the model achieves an AUC of 73.90\%, 76.52\% and 75.54\% on DFDC, DFD and CelebDF respectively, as shown in Table~\ref{Table7}. The addition of behavioral features improves 12.90\%, 10.86\% and 10.89\% AUC on DFDC, DFD and CelebDF. The AUC gains for final combination of features on cross-dataset evaluations are 5.25\%, 4.24\% and 3.29\%.  Our proposed combination of features achieves the best generalization towards new types of deepfakes and expressions swaps. 


\begin{table}
  \centering
\caption{Effectiveness of different combination of features on cross-dataset evaluation. Model is trained on FF++ and tested on DFDC, DFD and CelebDF.
}
\begin{tabular}{l c c c} 
 \hline
 Features               &  DFDC & DFD & CelebDF     \\
 \hline
 DIF                    &  73.90  &  76.52 & 75.54    \\
 \hline
 DIF+B                  &  86.80  &  87.38 & 86.43    \\
 \hline
 DIF+B+G                &  92.05  &  91.62 & 89.72     \\
 \hline
\end{tabular}

\label{Table7}
\end{table}


\section{Conclusion and Future Direction}


To address the limitation of poor generalizability in deepfake detectors, this paper presents DBaGNet; a triplet loss based framework with improved generalization across different manipulation. First, a DBaG descriptor is designed that is composed of identity, behavior, and geometric cues for a holistic analysis of given media. The combination of DBaG includes deep features that cover the identity of the subject, behavior analysis using blend shape features, and facial geometry analysis based on distances among various facial parts. After extracting the DBaG, the obtained features are forwarded to the proposed DBaGNet for representation learning, followed by a triplet loss-based classifier where distance-based classification is performed. Extensive experimentation is carried out on six large-scale datasets including cross-manipulation and cross-dataset evaluation which demonstrate superiority of the proposed framework against state-of-the-art methods. In the future, our aim is to explore some biological cues such as emotions or gaze estimation to strengthen the DBaG descriptor. 

\bibliographystyle{IEEEtran}
\bibliography{references}
\vspace{-1cm}
\begin{IEEEbiography}
[{\includegraphics[width=1.0in,height=1.75in,clip,keepaspectratio]{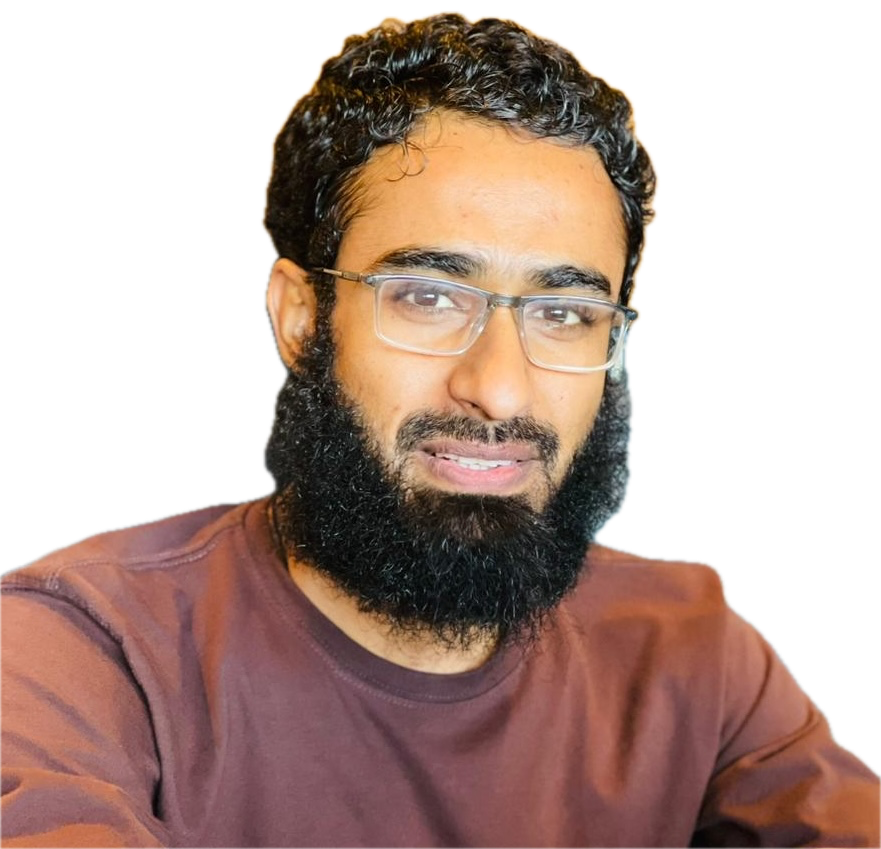}}]{Muhammad Umar Farooq}
is currently pursuing a Doctor of Philosophy in Computer Science at the School of Engineering and Computer Science, Oakland University, Rochester, Michigan, USA. His research interests include Digital Forensics with a particular focus on Deepfakes generation and detection, Explainable AI, and Neuro-symbolic AI. He aims to enhance the trustworthiness of digital media and to combat the negative implications of generative AI.
\end{IEEEbiography}

\vspace{-1cm}

\begin{IEEEbiography}[{\includegraphics[width=1in,height=1.75in,clip,keepaspectratio]{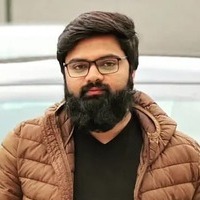}}]{Awais khan}
is currently pursuing a Doctor of Philosophy in Computer Science at the School of Engineering and Computer Science, Oakland University, Rochester, Michigan, USA. His research interests revolve around the captivating realm of speech processing, with a particular emphasis on topics like Deepfake Detection, Automatic Speaker Verification, and Anti-Spoofing. He aims to enhance security and combat emerging threats within voice-based technologies.
\end{IEEEbiography}

\vspace{-1cm}
\begin{IEEEbiography}[{\includegraphics[width=1in,height=1.75in,clip,keepaspectratio]{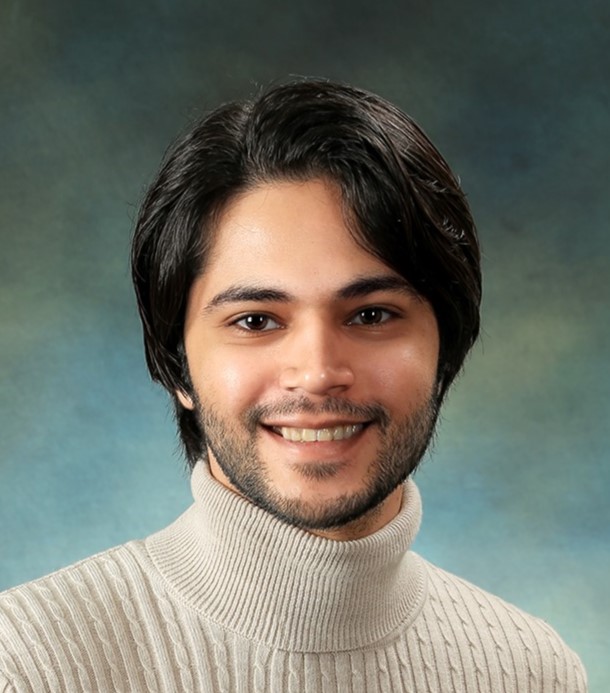}}]{Ijaz ul Haq} received the bachelor’s degree in computer science from
Islamia College Peshawar, Pakistan, in 2016 and Ph.D. degree in digital contents from Sejong University, Seoul, South Korea in 2022.
He is currently working as Research Fellow in University of Michigan. His research interests include the multimodal and  neurosymbolic learning for multimedia forensics. He has published several papers in reputed peer reviewed international journals and conferences including IEEE TII, Information Fusion, and Future Generation Computer Systems. 
\end{IEEEbiography}

\vspace{-1cm}

\begin{IEEEbiography}[{\includegraphics[width=1in,height=1.75in,clip,keepaspectratio]{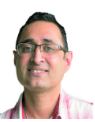}}]{Khalid Mahmood Malik} (SMIEEE) is currently a Professor of Computer Science and director of cybersecurity, at University of Michigan-Flint. His current research interests include use of multimodal and neurosymbolic AI for multimedia forensics, web security, and clinical management of cerebrovascular diseases. He also works on trustworthy AI, secure multicast protocols for intelligent transportation systems, and automated knowledge graph generation using large language models. His research is supported by multiple National Science Foundation awards, Department of Energy, Department of Defense, Michigan Translational Research \& Commercialization and Brain Aneurysm Foundation.

\end{IEEEbiography}






\end{document}